\documentclass[letterpaper]{article} 
\usepackage{aaai24}  
\usepackage{times}  
\usepackage{helvet}  
\usepackage{courier}  
\usepackage[hyphens]{url}  
\usepackage{graphicx} 
\usepackage{natbib}  
\usepackage{caption} 
\usepackage{algorithm}
\usepackage{algorithmic}

\usepackage{newfloat}
\usepackage{listings}
\DeclareCaptionStyle{ruled}{labelfont=normalfont,labelsep=colon,strut=off} 
\floatstyle{ruled}
\newfloat{listing}{tb}{lst}{}
\floatname{listing}{Listing}
\title{Deep Fusion: Capturing Dependencies in Contrastive Learning via Transformer Projection Heads}

\author{%
Huanran Li \textsuperscript{\rm 1}, Daniel Pimentel-Alarc\'on \textsuperscript{\rm 2}}
\affiliations{
\textsuperscript{\rm 1}Department of Electrical Engineering\\
\textsuperscript{\rm 2}Department of Biostatistics\\
\textsuperscript{\rm 1}\textsuperscript{\rm 2}Wisconsin Institute of Discovery\\
University of Wisconsin-Madison\\
\texttt{\{hli488\}\{pimentelalar\}@wisc.edu}
}

\usepackage{bibentry}

\usepackage{xcolor}

\definecolor{cvprblue}{rgb}{0.21,0.49,0.74}
\usepackage{enumitem}

\usepackage{todonotes}
\usepackage{url}
\usepackage{multirow} 
\usepackage{rotating} 

\usepackage{array} 

\usepackage{graphicx} 
\usepackage{caption}

\usepackage{amsmath}

\usepackage{amssymb} 

\usepackage{algorithm}
\usepackage{algorithmic}
\usepackage{multirow}
\usepackage{xcolor} 
\definecolor{shadecolor}{rgb}{.95,.95,.95}  

\usepackage{graphicx} 
\usepackage{amsmath}
\usepackage{amsfonts}
\usepackage{bm}
\usepackage{amsthm}
\usepackage{framed} 
\usepackage{enumitem}
\usepackage{listings}
\usepackage{color} 
\definecolor{codegreen}{rgb}{0,0.6,0}
\definecolor{codegray}{rgb}{0.5,0.5,0.5}
\definecolor{codepurple}{rgb}{0.58,0,0.82}
\definecolor{backcolour}{rgb}{0.95,0.95,0.92}

\lstdefinestyle{mystyle}{
    backgroundcolor=\color{backcolour},
    commentstyle=\color{codegreen},
    keywordstyle=\color{magenta},
    numberstyle=\tiny\color{codegray},
    stringstyle=\color{codepurple},
    basicstyle=\ttfamily\footnotesize,
    breakatwhitespace=false,         
    breaklines=true,                 
    captionpos=b,                    
    keepspaces=true,                 
    numbers=left,                    
    numbersep=5pt,                  
    showspaces=false,                
    showstringspaces=false,
    showtabs=false,                  
    tabsize=2,
    language=Python
}

\theoremstyle{plain}
\newtheorem{theorem}{Theorem}

\newtheorem{definition}{Definition}

\theoremstyle{remark}

\usepackage{xcolor}         

\def \U{\mathcal{U}}

\def \bu{\mathbf{u}}

\def \D{\mathcal{D}}

\def \S{\mathcal{S}}
\def \tX{\tilde{\mathbf{X}}}
\def \tx{\tilde{\mathbf{x}}}

\def \X{\mathbf{X}}
\def \W{\mathbf{W}}
\def \V{\mathbf{V}}
\def \K{\mathbf{K}}
\def \Q{\mathbf{Q}}
\def \A{\mathbf{A}}

\def \x{\mathbf{x}}

\def \sk{\mathsf{k}}
\def \z {\mathbf{z}}

\begin{document}
\maketitle

\begin{abstract}
Contrastive Learning (CL) has emerged as a powerful method for training feature extraction models using unlabeled data. Recent studies suggest that incorporating a linear projection head post-backbone significantly enhances model performance. In this work, we investigate the use of a transformer model as a projection head within the CL framework, aiming to exploit the transformer's capacity for capturing long-range dependencies across embeddings to further improve performance. Our key contributions are fourfold: First, we introduce a novel application of transformers in the projection head role for contrastive learning, marking the first endeavor of its kind. Second, our experiments reveal a compelling "Deep Fusion" phenomenon where the attention mechanism progressively captures the correct relational dependencies among samples from the same class in deeper layers. Third, we provide a theoretical framework that explains and supports this "Deep Fusion" behavior. Finally, we demonstrate through experimental results that our model achieves superior performance compared to the existing approach of using a feed-forward layer.
\end{abstract}

\section{Introduction}




Contrastive Learning (CL) has recently garnered significant attention due to its effectiveness in training feature extraction models without the need for labeled data. Along this trajectory, several renowned models have been introduced, including SimCLR\cite{chen2020simple}, Momentum Contrast (MoCo) \cite{he2020momentum}, Contrastive Multiview Coding (CMC) \cite{tian2020contrastive}, VICReg\cite{bardes2021vicreg}, BarLowTwins\cite{zbontar2021barlow}, and \cite{wu2018unsupervised, henaff2020data}.
These approaches share a common framework: during training, the objective is to minimize the distance between augmented versions of images from the same source while simultaneously maximizing the distance between images from different sources.
Following the training phase, the model is commonly combined with a feed-forward neural (FFN) decoder to fine-tune its performance using labeled data. Empirical evidence demonstrates that these models can achieve performance levels comparable to fully-supervised models, even when trained with a relatively limited amount of labels (approximately $10\%$) on moderate to large datasets \cite{jaiswal2020survey}. 

 The core concept of contrastive learning hinges on training models by comparing the cosine similarities of embeddings from either similar or dissimilar images. This method involves closely aligning embeddings from identical images by maximizing their cosine similarity, while distancing those from different ones by minimizing it. This process enables the model to effectively learn critical image features. Interestingly, this approach shares similarities with the attention mechanism of Transformer models, which also measures the pairwise cosine similarity of embeddings from the key and query feed-forward layers, but combines them through a similarity-weighted sum operation. Additionally, research highlights that using a non-linear projection head, typically comprising feed-forward layers, significantly boosts the effectiveness of contrastive learning \cite{chen2020simple, chen2020improved, xiao2020should, you2021graph, wang2023adaptive, zheng2021weakly, wang2021dense}. Therefore, replacing the projection head with a more expressive architecture like a Transformer, and leveraging the Transformer's ability to detect long-range dependencies among embeddings in the projection head, could be a promising avenue to further enhance the performance of contrastive learning. This would harness the strengths of both the contrastive framework and the Transformer for more robust feature extraction.

\begin{figure*}
    \centering
    \includegraphics[width=\textwidth, trim={0 5.9cm 1.3cm 0}, clip]{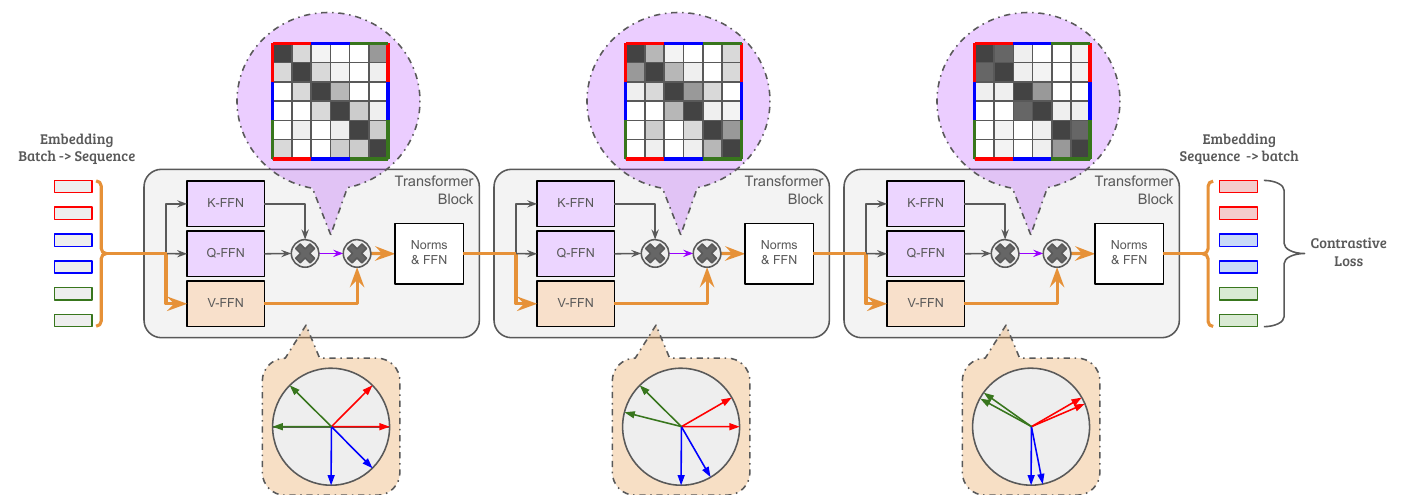}
    \caption{Deep Fusion within the Transformer Projection Head for Contrastive Learning. Embeddings from the backbone network are transformed into a sequence. As the process unfolds, the attention mechanism progressively identifies and amplifies relational dependencies among samples of the same class in deeper layers. This indicates an unsupervised 'fusion' of samples, drawing them closer to each other based on class similarity, without the need for explicit label supervision.}
    \label{fig:attention_map}
\end{figure*}

\textbf{Deep Fusion.} In this paper, we investigate the effects of incorporating a Transformer projection head into the SimCLR framework for contrastive learning. Our approach involves converting a batch of image embeddings from the backbone network into a sequence, which is then used to train the Transformer projection head using a contrastive loss function. A significant outcome of our experiments is the identification of a phenomenon we term "Deep Fusion." This phenomenon is characterized by the attention mechanism's increased ability to accurately identify relational dependencies among samples of the same class in deeper layers, as illustrated in Figure \ref{fig:attention_map}. Following each attention mechanism, the similarity-weighted sum operation moves the embeddings closer to their positive pairs, a process we refer to as "Fusion". As a result, the embeddings, after being enhanced by multiple Transformer layers, form a more clustered structure.

\textbf{Theory.} We further develop a theoretical framework that explains and supports the "Deep Fusion" phenomenon. Our analysis reveals that "Deep Fusion" occurs when the attention mechanism’s key and query weights effectively identify the underlying subspace structure of each class. Specifically, when key and query weights find vectors that are orthogonal to each cluster's subspace, the layer actively enhances the structure of the attention matrix to better reveal the underlying clusters. Our theoretical bounds show that the efficacy of this fusion mechanism is improved by lower noise levels within each cluster, larger spatial separation between different clusters, and shorter input sequences.

\textbf{Experiment.}
Finally, our empirical findings demonstrate that our model, which incorporates a Transformer projection head, significantly outperforms traditional contrastive learning methods that utilize a feed-forward layer. This superior performance is consistently observed across various datasets, including CIFAR10, CIFAR100, and ImageNet-200, and under different configurations of batch sizes and model dimensions.

\textbf{Paper Organization.} We first describe the contrastive learning framework SimCLR and detail our modifications incorporating a Transformer projection head. Then, we provides a comprehensive theoretical analysis of the Deep Fusion phenomenon observed in our model. Following that, we review the relevant literature on self-supervised learning and Transformers, placing our work within the broader research context. Finally, we presents the experimental results.

\section{Transformer Projection Head in Contrastive Learning}
\label{sec-model}

 In this section, we initially review the contrastive learning framework employed in this study, namely Simple Framework for Contrastive Learning of Representations (SimCLR) \cite{chen2020simple}. Subsequently, we describe the process of integrating Transformer models as projection heads within the SimCLR framework. Lastly, we present the "Deep Fusion" phenomenon, illustrating how Transformers progressively discern relationships among samples from identical classes through self-supervised learning, without the need for training labels.

SimCLR consists of four main components: a base encoder, a projection head, augmentation strategies, and contrastive loss. The base encoder, which is typically a convolutional neural network (CNN) or a vision transformer (ViT) \cite{dosovitskiy2020image}, maps an augmented image to a representation vector. This vector is then transformed by the Projection Head, a small neural network, into a projection that is used for calculating the contrastive loss, thereby focusing on features significant for contrastive learning. During the training process, positive pairs are created by applying two random augmentations to the same image. The augmentation strategy contains color jitter, random crop, random gray scale, gaussian blur, flip, and rotation. The output of the model contains a batch of embedding vectors (denote $\z_i$), and the NT-Xent Loss \cite{chen2020simple} (normalized temperature-scaled cross entropy loss) is calculated by:
\begin{align}
    \ell(i, j) = -\log \frac{\exp\left(\text{sim}(z_i, z_j) / \tau\right)}{\sum_{k=1}^{2N} 1_{k \neq i} \exp\left(\text{sim}(z_i, z_k) / \tau\right)}
    \label{contrast_loss}
\end{align}
where $x_i$ and $x_j$ are two augmented views of the same image; $z_i = f(x_i)$ and $z_j = f(x_j)$ are the vectors obtained from the base encoder and projection head. The function 
$
\text{sim}(u, v) = \frac{u \cdot v}{\|u\| \|v\|}
$
calculates the cosine similarity, and $\tau$ represents a temperature scaling parameter. 

 In the context of self-supervised learning, samples from the same class are treated as negative samples due to the absence of labels. If the projection head is overly simplistic, the contrastive loss might adversely impact the learning process of the backbone network by excessively penalizing the similarities of samples from the same class \cite{wang2023adaptive}. In contrast, Transformer models, known for their high expressiveness and ability to uncover latent dependencies across sequences, are well-suited for the objectives of contrastive learning.
 
\textbf{Transformer Projection Head.}
To switch the projection head to a transformer model, we convert the batch of embeddings from the base encoder into a sequence for the input of the Transformer projection head. Following \cite{vaswani2017attention}, each Transformer decoder block has a Multi-Head Attention layer defined as:
    \[
    \text{MultiHead}(\Q, \K, \V) = \text{Concat}(\text{head}_1, \dots, \text{head}_h)\W^O
    \]
    \[
    \text{head}_i = \text{softmax}\left(\frac{(\Q\W_i^Q)(\K\W_i^K)^T}{\sqrt{d_k}}\right) \V\W_i^V
    \]
    where $\W^O, \W_i^Q, \W_i^K, \W_i^V$ are all learnable weights, $d_k$ is the dimension of the keys, and $\K,\Q,\V$ are the same as input.
 Each block also includes a fully connected feedforward network with GeLu and layer Norm, which is applied to each position separately and identically. The output sequence is then used as the final output embeddings for self-supervised training and inference.

\section{Deep Fusion: Grouping Inputs by Attention without Supervision}
\label{sec-theory}

During our experiments, we observed an intriguing phenomenon we have termed "Deep Fusion," where the attention maps begin to discern the true class relationships between samples, independent of labels in training. This capability to recognize true labels improved with the depth of the layers within the transformer projection head, suggesting that the transformer autonomously groups the samples without supervision. To demonstrate Deep Fusion, we trained a 4-layer CNN followed by a 10-layer Transformer projector on the MNIST dataset for 200 epochs. We used a learning rate of 0.1 with a cosine annealing scheduler, SGD optimizer, and a batch size of 1024. During inference, we visualized the attention weights for layers 1, 3, 6, 8, and 10, arranged sequentially from left to right in Table \ref{tab:CIFAR10}. We noted that as the layers became deeper, the attention maps increasingly resembled a block-diagonal adjacency matrix of the input, even without labeled training data.

\begin{table*}
    \centering
    \setlength{\tabcolsep}{1pt} 
    \begin{tabular}{ccccc} \hline
        Layer 1 & Layer 3 & Layer 6 & Layer 8 & Layer 10 \\
        \includegraphics[width=0.195\textwidth, trim = {4.3cm 3.0cm 9cm 3.27cm}, clip]{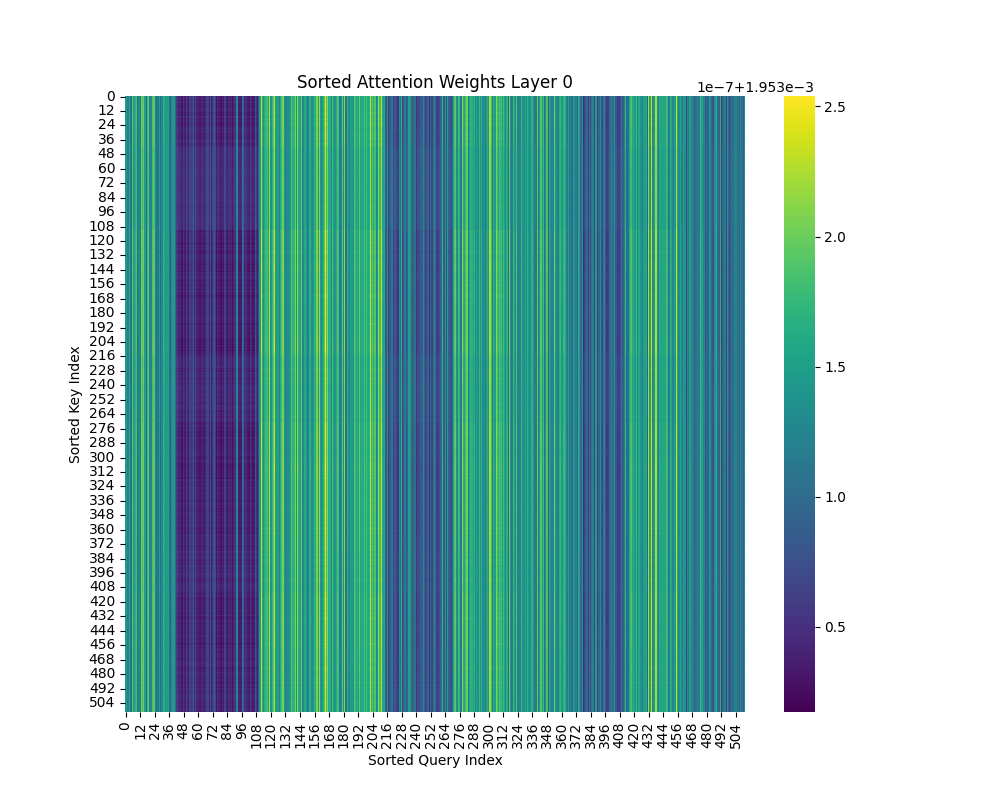} &
        \includegraphics[width=0.195\textwidth, trim = {4.3cm 3.0cm 9cm 3.27cm}, clip]{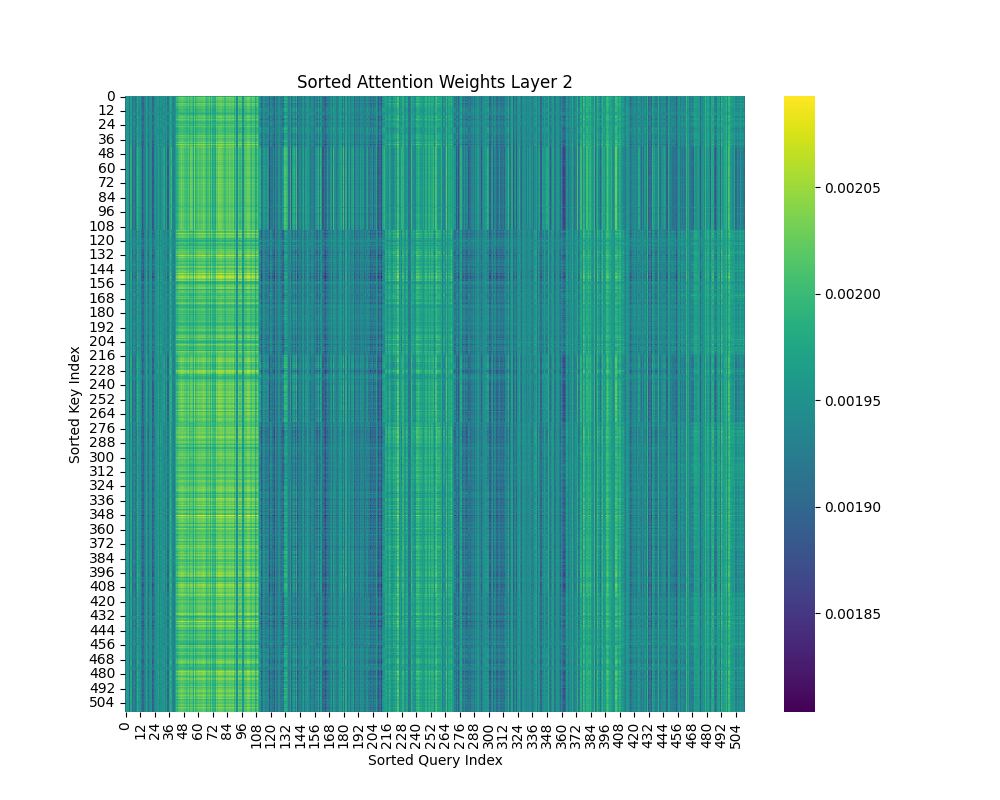} &
        \includegraphics[width=0.195\textwidth, trim = {4.3cm 3.0cm 9cm 3.27cm}, clip]{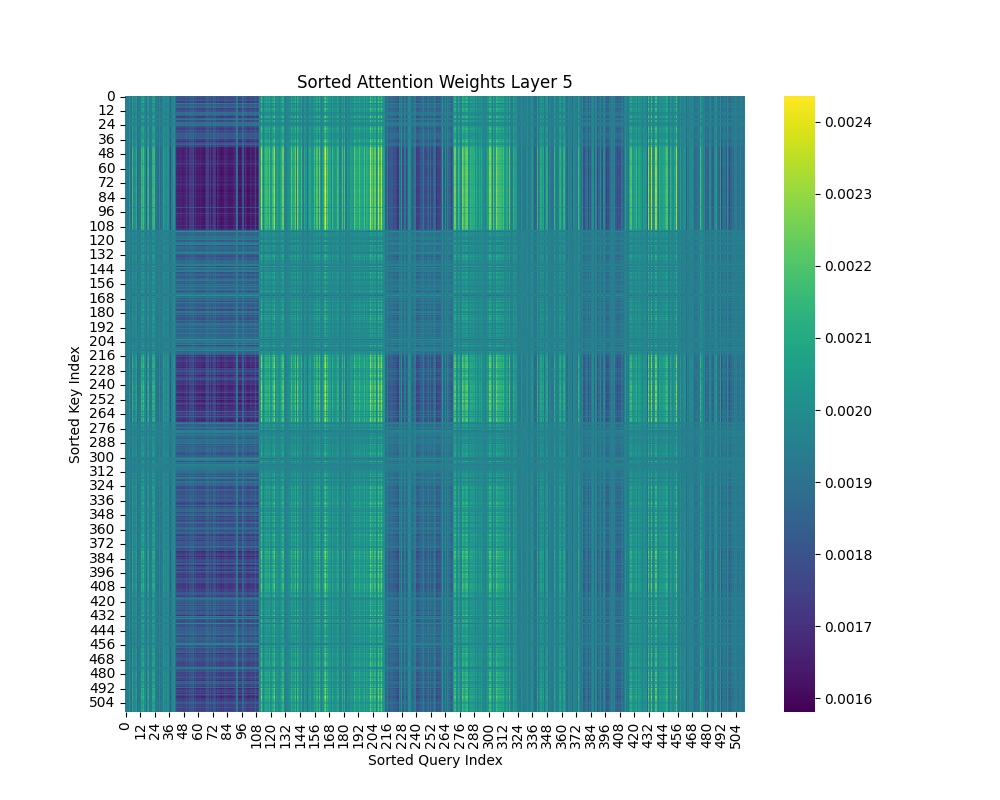} &
        \includegraphics[width=0.195\textwidth, trim = {4.3cm 3.0cm 9cm 3.27cm}, clip]{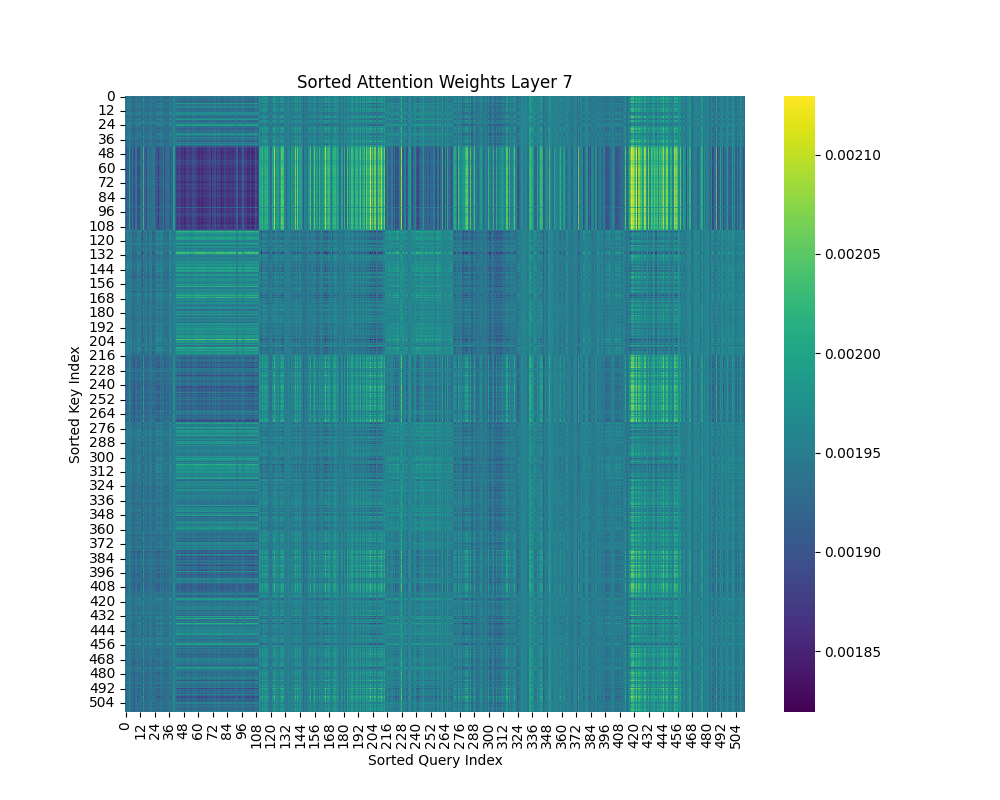} &
        \includegraphics[width=0.195\textwidth, trim = {4.3cm 3.0cm 9cm 3.27cm}, clip]{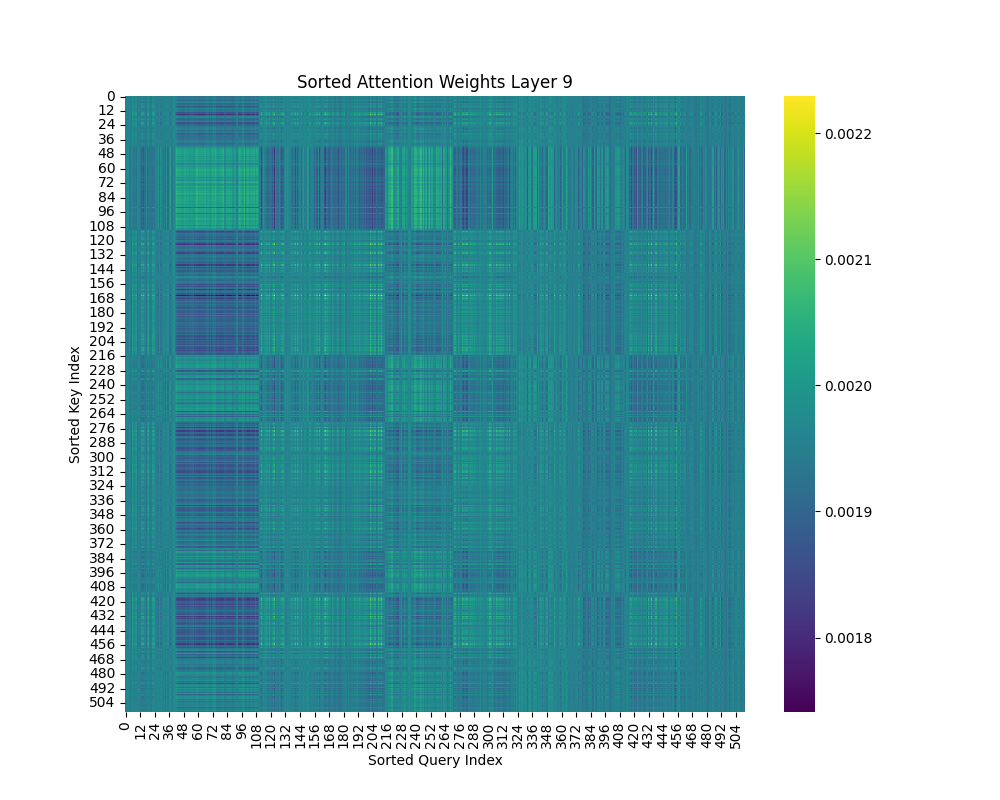}  \\\hline
    \end{tabular} \vspace{0.1cm}
    \caption{Attention weights of the first head across layers 1, 3, 6, 8, and 10 of the Transformer projector. Training setup for the MNIST dataset involved a 4-layer CNN followed by a 10-layer, 4-head Transformer projector, trained over 200 epochs. A learning rate of 0.1 was employed alongside a cosine annealing scheduler and SGD optimizer, with a batch size of 1024.}
    \label{tab:CIFAR10}
\end{table*}

In the remainder of this section, we will theoretically analyze the deep fusion effect, demonstrating that this phenomenon arises when the attention mechanism's key and query weights, $\W^Q, \W^K$ effectively capture the underlying subspace structure of each class simultaneously. Specifically, we will initially examine a noiseless scenario to illustrate that if the input data matrix $\mathbf{X}$ displays a structure of subspace clusters, then a single-layer Transformer model can accurately generate an attention matrix that differentiates samples within the same subspace.

In the latter part of this section, we extend our findings to demonstrate that even with noise present, each layer of the Transformer effectively amalgamates samples from the same class, exemplifying the true Deep Fusion phenomenon. Our theoretical analysis suggests that the results of contrastive learning are significantly impacted by factors such as the level of noise, the spatial separation between clusters, and the length of the sequence.

\subsection{One-Layer Fusion without Noise}
We consider the data matrix $\mathbf{X}$ to consist of multiple low-dimensional subspaces, represented as $\{\U_1, \ldots, \U_\mathcal{K}\}$, where each row of $\mathbf{X}$ is part of one of these subspaces, with the rank of each subspace not exceeding $r$. We assume that  no single subspace is linearly dependent on a combination of all other subspaces, i.e. $(\mathcal{K}-1) r < m$, where $m$ denotes ambient dimension and maximum rank of $\X$.
For each low-rank subspace $\mathcal{U}_k$, there always exists a unit vector $\mathbf{u}^\perp_k$ that is perpendicular to $\mathcal{U}_k$, such that for any vector $\mathbf{x}_i$ not belonging to $\mathcal{U}_k$, the cosine similarity between $\mathbf{x}_i$ and $\mathbf{u}^\perp_k$ is non-zero. Furthermore, we introduce \textit{Cluster Integrity} as a measurement of the lower bound of the maximum cosine-similarity.
\begin{definition}
   Given a collection of input samples $\X$, where each sample $\mathbf{x}_i$ belongs to one of the subspaces among $\{\mathcal{U}_1, \ldots, \mathcal{U}_\mathcal{K}\}$, the \textbf{Cluster Integrity} $\rho$ is defined as the lower bound of $\{\rho_1, \ldots, \rho_\mathcal{K}\}$, where
\begin{align}
    \rho_k := \max_{\mathbf{u} \perp \mathcal{U}_k } \min_{\mathbf{x}_i \notin \mathcal{U}_k} |\mathbf{x}_i^\top \mathbf{u}|
    \label{eq:projection-lb}
\end{align}
\label{class-int}
\end{definition}
In words, \textit{Cluster Integrity}, ranging from 0 to 1, measures the degree of separation between subspace clusters. If two clusters $\mathcal{U}_a$ and $\mathcal{U}_b$ are well-separated, then a vector $\mathbf{u}_{a}^{\perp}$, orthogonal to $\mathcal{U}_a$, can be found for large cosine-similarity of all points in $\mathcal{U}_b$, resulting in a higher $\rho$ value. Conversely, if two clusters $\mathcal{U}_a$ and $\mathcal{U}_b$ are not well-separated, a vector $\mathbf{u}^{\perp}_a$ orthogonal to $\mathcal{U}_a$ would likely also be orthogonal to $\mathcal{U}_b$, leading to small similarity and a lower $\rho$ value.
The motivation for employing \textit{Cluster Integrity} as a metric is to maintain consistency with the attention mechanism, which calculates the attention matrix using cosine similarity between all pairs of transformed vectors $\mathbf{x}_i$ and $\mathbf{x}_j$. For the particular dataset, we can use an algorithm to estimate clustering integrity through a greedy search, as detailed in Appendix \ref{algo-search-rho}.

Our first theoretical result, summarized in the following theorem, demonstrates that if every sample in the matrix $\X$ originates from one of the defined subspaces, then the resulting attention matrix $\A$ can be theoretically constrained. Under this constraint, samples that belong to the same class will exhibit a cosine similarity that is always greater than a polynomial function of $\rho$; and samples from different classes are predicted to have a cosine similarity of 0.

\begin{framed}
\begin{theorem}
\label{existence}
Given a collection of input samples $\X$, where every sample $\x_i$ comes from one subspaces among $\{\mathcal{U}_1, ..., \mathcal{U}_\mathcal{K}\}$, there always exists a pair of parameters $(\W^{Q*}, \W^{K*})$ such that the attention matrix $\A$ calculated by $\A:= (\Q\W^{Q*})(\K\W^{K*})^\top$ has a block-diagonal form, ie.
$$
\begin{cases}
[\A]_{i,j} = 0,  & \ \ \ \x_i, \x_j \text{ in different clusters} \\
[\A]_{i,j} > \nu_i \rho^2,  & \ \ \ \x_i, \x_j \text{ in the same cluster}
\end{cases}
$$
where $\nu_i$ is the number of samples in same cluster as $\x_i$, and $\rho$ is the \textit{Class Integrity} defined in Definition \ref{class-int}.
\end{theorem}
\end{framed}


In words, Theorem \ref{existence} shows that the linear transformation of $\x_i$ and $\x_j$ has a cosine similarity of zero if they are from different clusters, and a positive similarity with lower bound if they belong to the same cluster. To illustrate this, we present a simple example. Let $\X$ be a matrix with $n = 4$ samples and an ambient dimension of $m = 4$.
Consider two rank-1 subspaces, $\{\mathcal{U}_1, \mathcal{U}_2\}$, where $\x_1, \x_2 \in \mathcal{U}_1$, and $\x_3, \x_4 \in \mathcal{U}_2$. The bases of these subspaces can be represented as $\{\bu_1,\bu_2\} \in \mathbb{R}^{4}$. Define:
$$\W^* = \begin{bmatrix}
    \bu_2^\perp & \bu_2^\perp & \bu_1^\perp & \bu_1^\perp 
\end{bmatrix},$$
where $\bu_1^\perp$ and $\bu_2^\perp$ are bases orthogonal to $\mathcal{U}_1$ and $\mathcal{U}_2$, respectively, i.e. $\bu_1^\top \bu_1^\perp = \bu_2^\top \bu_2^\perp = 0.$
Subsequently, applying the linear transformation to matrix $\mathbf{X}$ using $\mathbf{W}^*$ will reveal the clustering information of the samples:
\begin{align}
\X\W^* = 
\begin{bmatrix}
    \x_1^\top \bu_2^\perp &\x_1^\top \bu_2^\perp & 0 & 0  \\
    \x_2^\top \bu_2^\perp &\x_2^\top \bu_2^\perp & 0 & 0   \\
     0 &0 &\x_3^\top \bu_1^\perp & \x_3^\top \bu_1^\perp \\
     0 &0 & \x_4^\top \bu_1^\perp & \x_4^\top \bu_1^\perp  \\
\end{bmatrix}
\end{align}
Thus, by setting $\W^{K*} = \W^{Q*} = \W^*$, we obtain the block-diagonal attention matrix $\A$,
where each non-zero entry in $\A$ is lower-bounded by $2\rho^2$.
What follows is a formal proof of Theorem \ref{existence}.

    \begin{proof}
    Consider a matrix $\mathbf{X} \in \mathbb{R}^{n \times m}$, representing $n$ samples each in an $m$-dimensional ambient space. These samples are partitioned into $C$ clusters. Let $c_i$ denote the cluster assignment for the sample $\mathbf{x}_i$, where $c_i \in \{1,2,...,C\}$.
    
    Introduce a weight matrix $\mathbf{W} \in \mathbb{R}^{m \times n}$, where each column $\mathbf{w}_i$ is constructed to be orthogonal to the subspaces corresponding to all clusters except the one containing $\mathbf{x}_i$.
    
    Consequently, the product $\mathbf{XW}$ yields:
    \[
    [\mathbf{XW}]_{i,j} = \mathbf{x}_i^\top \mathbf{w}_j = \begin{cases} 0 & \text{if } c_i \neq c_j, \\
     \geq \rho & \text{if } c_i = c_j,
    \end{cases}
\]
where $\rho$ is a predefined lower bound.

For any pair of samples $\mathbf{x}_i$, $\mathbf{x}_j$ in the same cluster, the $i$-th and $j$-th rows of $\mathbf{XW}$, denoted as $[\mathbf{XW}]_{i,\cdot}$ and $[\mathbf{XW}]_{j,\cdot}$ respectively, have coinciding non-zero entry locations. Therefore,
\begin{align}
    [\mathbf{A}]_{i,j} = [\mathbf{XW}]_{i,\cdot}^\top [\mathbf{XW}]_{j,\cdot} \geq \nu_i \rho^2,
\end{align}
where $\nu_i$ is the number of samples in the same cluster as $\mathbf{x}_i$.

Conversely, for any pair of samples $\mathbf{x}_i$, $\mathbf{x}_j$ belonging to different clusters, the non-zero entry locations in the $i$-th and $j$-th rows of $\mathbf{XW}$ are entirely distinct. Thus,
\begin{align}
    [\mathbf{A}]_{i,j} = [\mathbf{XW}]_{i,\cdot}^\top [\mathbf{XW}]_{j,\cdot} = 0.
\end{align}
\end{proof}

\subsection{Deep Fusion}

To address the presence of noise, the clear block-diagonal structure may not be immediately apparent within the first block. Therefore, our objective is to progressively enhance the block-diagonal nature of $\A^{\ell}$ with each subsequent layer $\ell$. By stacking a sufficient number of layers, the final layer's output $\A^d$ should possess the desired property to effectively recognizing clusters.

To achieve this objective, our initial step involves bounding the magnitude of noise. Let $\tilde{\mathbf{X}}$ denote the input data with noise such that each sample (row) of $\tilde{\mathbf{X}}$ has a non-negative cosine similarity with the corresponding sample in $\mathbf{X}$.
And the cosine similarity is lower-bounded by a universal constant $\varepsilon \in [0,1]$:
\begin{align}
\mathbf{x}^\top_i \tilde{\mathbf{x}}_i \geq (1-\varepsilon)
\label{eq:noise-bound}
\end{align}
With this, we can further bound the projection of $\tx_i$ onto the orthogonal complement $\bu^\perp_\sk$:\\
    1) When $\x_i$ belongs to $\U_\sk$,
            \begin{align*}
                \tx_j^\top \bu^\perp_\sk \leq \sqrt{(1-(1-\varepsilon)^2)} =: \delta
            \end{align*} 
    2) When $\x_i$ does NOT belong to $\U_\sk$,
            \begin{align*}
                \tx_j^\top \bu^\perp_\sk \geq (1-\varepsilon)\rho - \sqrt{(1-(1-\varepsilon)^2)(1-\rho^2)} =: \Delta
            \end{align*}
Intuitively, $\delta$ represents the similarity upperbound between the input in class $\sk$ and its perpendicular vectors, and $\Delta$ represents the similarity lowerbound between the input in class $\sk$ and perpendicular vectors to other class. 

To validate the enhancement of cluster structure within the attention matrix at each layer, we introduce a metric specifically devised to evaluate the similarity between matrix $\mathbf{A}$ and an ideal block-diagonal matrix. This metric evaluates the proportion of the lowest attention value among samples within the same class to the highest attention value among samples from different classes. If our matrix $\A$ closely resembles a block-diagonal form, this ratio is expected to approach $+\infty$, indicating a high level of class-specific clustering within the matrix. 

\begin{definition}
\label{sharp-def}
The \textbf{sharpness} of an attention matrix $\A$ is defined as the infimum of the ratio between the attention of two points within the same cluster and the attention of two points belonging to different clusters, i.e.,
$$\S(\A) := \inf_{i,j,p,q} \frac{[\A]_{i,j}}{[\A]_{p,q}},$$
where $\x_i, \x_j$ are from the same cluster and $\x_p, \x_q$ are from different clusters.
\end{definition}

In words, \textit{sharpness} is quantifying how well-defined are our clusters. It bears similarity to the Silhouette index (SI) \cite{rousseeuw1987silhouettes}, commonly used for assessing cluster separation. However, a key distinction lies in the measurement approach: while SI calculates cluster quality using average distances for both intra-cluster and inter-cluster sample pairs, the \textit{sharpness} metric evaluates the ratio of the maximum intra-cluster distance to the minimum inter-cluster distance, and thus, imposes more stringent requirements on the structural integrity of the clusters.

Building upon this foundation, we substantiate that, mirroring the configuration of noise-free data, every Transformer block possesses the capacity to enhance the \textbf{sharpness} of its attention matrix compared to the preceding layer. This enhancement is quantifiable by a constant factor determined by $\delta, \Delta$, and batch size $n$.
\begin{framed}
\begin{theorem}
Given a input matrix $\tX$, where each sample $\tx_i$ lies near one of the subspaces among $\{\mathcal{U}_1, ..., \mathcal{U}_\mathcal{K}\}$ with noise bounded by \eqref{eq:noise-bound}, there always exists a pair of parameter $(\W^{Q*}, \W^{K*})$ such that each Transformer block increases the \textbf{sharpness} of its similarity matrix $\A^\ell$ by at least a factor of $O\left(\frac{1}{n} e^{n(\delta^2 - \Delta^2) + \delta}\right)$.
\label{main-theorem}
\end{theorem}
\end{framed}

Theorem \ref{main-theorem} states that under the assumption of separable clusters ($\delta < \Delta$), the sharpness improvement ratio over layers decays as the sequence lengthens. This implies that the Deep Fusion effect is more obvious with smaller noise, larger inter-cluster distance, and shorter sequence.
The proof of Theorem \ref{main-theorem} follows by the same arguments as that of Theorem \ref{existence}, except the off-diagonal blocks are upper bounded by $\delta$ instead of zero, and the diagonal blocks are lower bounded by $\Delta$. The detailed proof can be found in the Appendix \ref{pf-T2}.

\section{Related Work}

\label{sec-relatedwork}
This section reviews two key areas of recent research relevant to this study. The first part explores the latest developments in contrastive learning, highlighting advancements over the projection head that enhance representation quality and model performance. The second part delves into the application of Transformer models in computer vision tasks, discussing how self-attention mechanisms could help improving model performce over classification, segmentation, generation tasks.

\begin{table*}[h]
    \centering
\setlength{\tabcolsep}{2pt} 
    \begin{tabular}{cc|cc|ccc}\hline
       Dataset  & Backbone  & \multicolumn{2}{c|}{Projection Head}  & Unsup. Test  &  Sup. Test & Time \\ 
         & & Size  & Type  &  Accuracy & Accuracy & Cost(h)
       \\\hline
        CIFAR10 & ResNet18  & FFN &1.9M & 73.49\%  &  80.88\% &4.17$\pm$0.04 \\
        CIFAR10 & ResNet18  & TF &1.9M  & \textbf{76.65\% (+3.16\%)} &\textbf{83.26\%(+2.38\%)} &\textbf{2.53$\pm$2.08}\\ \hline
        CIFAR10 & ResNet18  & FFN &7.4M & 75.73\%  &  81.21\% &4.95$\pm$0.17 \\
        CIFAR10 & ResNet18  & TF &7.6M & \textbf{78.37\% (+2.64\%)} & \textbf{84.54\%(+3.33\%)} &\textbf{3.39$\pm$1.97}\\ \hline
        CIFAR100 & ResNet18 & FFN &1.9M & 37.42\% & 42.51\% &4.22$\pm$0.18 \\
        CIFAR100 & ResNet18  & TF &1.9M & \textbf{40.19\% (+2.77\%)} & \textbf{46.12\%(+3.61\%)}  &\textbf{ 2.98$\pm$1.88} \\\hline
       
            CIFAR100 & ResNet18  & FFN &7.4M & 38.1\% & 44.46\% & 4.61$\pm$0.16\\
    CIFAR100 & ResNet18  & TF &7.6M &\textbf{41.3\% (+3.20\%)} & \textbf{47.92\%(+3.46\%)} &\textbf{3.80$\pm$2.16} \\\hline
    ImageNet-200 & ResNet50  & FFN  &1.9M & 20.57\% & 22.36\% & (2x)\textbf{3.05$\pm$0.09}\\
    ImageNet-200 & ResNet50  & TF  &1.9M & \textbf{22.82\% (+2.25\%)} & \textbf{24.89\% (+2.53\%)} & (2x){3.66$\pm$2.49}\\\hline
    ImageNet-200 & ResNet50  & FFN  &7.4M & 21.46\% & 23.05\% & (2x)\textbf{4.62$\pm$0.11} \\
    ImageNet-200 & ResNet50  & TF  &7.6M & \textbf{24.13\% (+2.67\%)} & \textbf{26.47\% (+3.42\%)} &(2x)6.21$\pm$4.22 \\ \hline

    \end{tabular}
    \caption{Performance of SimCLR with different projection head configurations. All models are trained for 1000 epochs with SGD on tuned learning rates, momentum, and weight decay.}
    \label{tab:Best}
\end{table*}

\subsection{Contrastive Learning}
In contrastive learning, each individual image is treated as a distinct class, and the model is trained to match various augmented versions of the same image amidst a backdrop of other images. A noteworthy milestone emerges in the form of a seminal study conducted by \cite{wu2018unsupervised}. This work introduces a non-parametric approach for gauging the similarity between features, thereby elevating accuracy levels across datasets like CIFAR10 and ImageNet. Another significant contribution comes from \cite{hjelm2018learning}, who showcase the potential of maximizing mutual information between an input and the output of a neural network encoder. 
Similarly, \cite{tian2020contrastive} build upon a comparable method to amplify the mutual information across diverse channels of a scene's view. The results of their experimentation corroborate the efficacy of this approach. Intriguingly, the utilization of contrastive learning has ushered in a remarkable shift in training larger networks with significantly fewer labeled data points, all the while achieving competitive outcomes on prominent datasets such as Imagenet \cite{chen2020simple} and PASCAL VOC \cite{henaff2020data}. A more recent contribution that stands out is the work undertaken by \cite{balestriero2022contrastive}. Remarkably, this study unveils the closed-form optimal representation and network parameters within the linear regime for prevalent self-supervised learning approaches, including VICReg \cite{bardes2021vicreg}, SimCLR \cite{chen2020simple}, and BarlowTwins \cite{zbontar2021barlow}. 

Incorporating a projection head into contrastive learning frameworks significantly enhances the quality of learned representations by directing the network to emphasize more informative and discriminative features. \cite{chen2020simple, chen2020improved, xiao2020should, wang2021dense} report that the addition of a projection head notably improves the performance in the linear evaluation protocol, which assesses the quality of learned representations by training a linear classifier on top of the frozen base network. \cite{you2021graph} introduce an augmentation-aware projection head mechanism that routes output features through different projection heads, corresponding to various augmentations selected at each training step, achieving performance comparable to or surpassing state-of-the-art methods. To address challenges in enforcing high similarity for positive pairs and low similarity for negative pairs, \cite{wang2023adaptive} propose using multiple projection heads, each generating distinct feature sets. Their adaptive multi-head contrastive learning (AMCL) method adjusts the similarity measures over pairs with individual adaptive temperatures, improving several contrastive learning models like SimCLR, MoCo, and Barlow Twins. Additionally, \cite{zheng2021weakly} propose a dual projection head framework where one head performs instance discrimination while the other employs a graph-based approach to generate weak labels for supervised contrastive learning, bringing similar images closer together.

\subsection{Transformer on Vision Tasks}

Transformers have significantly impacted vision tasks by employing self-attention mechanisms to model dependencies among different regions of the image. The work by \cite{vaswani2017attention} introduced the Transformer model, which eliminated the need for recurrent structures in sequence-to-sequence tasks and formed the foundation for vision applications. Extending this concept, \cite{parmar2018image} presented the Image Transformer, showcasing the adaptability of transformers to image generation. The Vision Transformer (ViT) \cite{dosovitskiy2020image} is the first to demonstrated that a standard Transformer applied to image patches could rival traditional convolutional networks in image recognition tasks. DETR \cite{carion2020end} simplified object detection by using transformers to directly predict bounding boxes and labels from image features. The Swin Transformer \cite{liu2021swin} introduced a hierarchical approach with shifted windows, enhancing scalability and efficiency for vision tasks. The Pyramid Vision Transformer (PVT) \cite{wang2021pyramid} utilized a pyramid structure to capture multi-scale features effectively, performing well across numerous vision benchmarks. Lastly, the Token-to-Token ViT (T2T-ViT) \cite{yuan2021tokens} proposed a progressive tokenization process, improving training efficiency and accuracy on image classification tasks. Collectively, these advancements emphasize the transformative role of transformers in computer vision.

\section{Experiment}
\label{sec-exp}

In this section, we showcase the experimental results of using a Transformer projection head on the CIFARs, and ImageNet-200 datasets. We compared the model's performance with an identical setup but with a feed-forward projection head within the SimCLR framework. Our results demonstrate that, across various configurations—including different backbones, batch sizes, and projection sizes—the Transformer projection head consistently outperforms the feed-forward projection head.

\textbf{Model Architecture.} We employed ResNet18 and ResNet50 as the backbones for CIFAR and ImageNet, respectively. The input dimension was set to 32x32, and the output dimension was 512. For the projection head, we configured three different sizes for both feed-forward and transformer layers:
The small projection head (2M parameters) comprises 5 feed-forward network (FFN) layers with a hidden dimension of 640, or 5 Transformer layers with 8-head attention and a feature dimension of 128.
The large projection head (7.5M parameters) comprises 7 FFN layers with a hidden dimension of 1024, or 5 Transformer layers with 16-head attention and a feature dimension of 256.

\textbf{Hyperparameters.} The batch size was fixed at $1024$, and the maximum number of epochs was set to $1000$. We used Stochastic Gradient Descent (SGD) with momentum of $0.9$ and a cosine annealing scheduler for weight decay of $0.001$. The learning rate were tuned on the validation set. The augmentation strategy follows the same setup in the \cite{chen2020simple}, and the temperature of the contrastive loss is fixed at $0.2$ (see Equation \eqref{contrast_loss}).

\textbf{Metrics.} We evaluated the performance of the methods by adhering to a common protocol. After the contrastive training phase, we extracted and froze all embeddings from the backbone (without projection head) and used 10\% of them to train a linear classifier consisting of a single-layer linear feed-forward layer followed by a softmax activation. We then reported the top-1 accuracy on the test set, referred to as "Supervised Test Accuracy." Additionally, we reported the "Unsupervised Test Accuracy" by extracting features from the test set and measuring the label agreement with the nearest neighbor (also in the test set) using Euclidean distance. Furthermore, we recorded the time taken for the entire training process, using one A100 GPU by default, with "(2x)" indicating that two GPUs were used during training.

\textbf{Results.} 
The results are shown in Table \ref{tab:Best}, which demonstrate that the Transformer projection head consistently outperforms the Feed-Forward projection head by $2.38\%$ to $5.26\%$ in supervised accuracy and $2.25\%$ to $3.20\%$ in unsupervised accuracy. We also observe that larger projection heads improve overall accuracy marginally, but at the cost of increased training time.

\textbf{Ablation Studies.} We conducted tests on CIFAR100 with various hyperparameter setups using ResNet18 and small projection heads, either feed-forward or Transformer. The default settings align with those in Table \ref{tab:Best}. 
\begin{itemize}[left=3pt, itemsep=0pt]
    \item \textbf{Batch Size.} We tested batch sizes of 512, 1024, 2048, and 4096, using identical setups with ResNet18 and a small projection head. The results are presented in Table \ref{tab:batch_size}.
    We measured the unsupervised accuracy and training time for both feed-forward (FFN) and Transformer (TF) projection heads. Our observations indicate that a batch size of 1024 yielded the highest unsupervised accuracy for both projection heads, with the Transformer projection head achieving 40.19\% and the feed-forward projection head reaching 35.43\%. The training time was also optimized at this batch size.
\begin{table}
    \centering
    \setlength{\tabcolsep}{2pt} 
    \begin{tabular}{c|c|cccc}
    \hline
        Proj. Head & Metric & 512 & 1024 & 2048 & 4096 \\ \hline
        \multirow{2}{*}{FFN} 
        & Unsup. Acc. & 34.87\% & \textbf{35.43\% }& 35.08\% & 35.02\% \\
        & Time(h) &\textbf{ 4.05 }& 4.22 & 5.78 & 5.38 \\ \hline
        \multirow{2}{*}{TF}
        & Unsup. Acc. & 38.54\% & \textbf{40.19\% }& 39.59\% & 35.12\% \\
        & Time(h) &\textbf{ 2.65} & 2.98 & 4.47 & 4.79\\ \hline
    \end{tabular}
    \caption{Ablation on different \textbf{batch sizes} on CIFAR100 with ResNet18 and small projection head.}
    \label{tab:batch_size}
\end{table}

    \item \textbf{Loss Temperature.} We experimented with different loss temperatures of $1.0$, $0.8$, $0.5$, $0.2$, $0.1$, $0.05$, and $0.02$ for equation \eqref{contrast_loss}. Our results (refer to Table \ref{tab:loss_temp}) indicate that both setups perform well with temperatures of $0.2$ and $0.1$, with $0.2$ being marginally better. The Transformer projection head achieving an unsupervised accuracy of 40.19\% and the feed-forward projection head reaching 35.43\%.
\begin{table}
    \centering
    \setlength{\tabcolsep}{2pt} 
    \begin{tabular}{c|cccc}
    \hline
        Proj. Head &  1 & 0.8 & 0.5 & 0.2 \\ \hline
        FFN &  13.21\% & 8.3\% & 18.43\% & \textbf{35.43\% } \\\hline
        TF & 4.25\% & 6.97\% & 11.08\% & \textbf{40.19\% } \\ \hline \hline
        Proj. Head  & 0.1 & 0.05 & 0.02 \\ \hline
        FFN & 33.95\% & 29.51\% &  26.65\%\\\hline
        TF &  34.09\% & 31.88\% & 1.18\% \\ \hline
    \end{tabular}
    \caption{Unsupervised Accuracy reported for ablation on different \textbf{loss temperatures} on CIFAR100 with ResNet18 and small projection head.}
    \label{tab:loss_temp}
\end{table}

    \item \textbf{Weight Decay.} We evaluated different weight decay rates of $10^{-1}$, $10^{-2}$, $10^{-3}$, and $10^{-4}$. Table \ref{tab:weight-decay} provides the results of on different weight decay rates using the CIFAR100 dataset with ResNet18 and a small projection head. We measured the unsupervised accuracy for both feed-forward (FFN) and Transformer (TF) projection heads. The findings suggest that a weight decay rate of $10^{-3}$ yields the best results, with the Transformer projection head achieving an unsupervised accuracy of 40.19\% and the feed-forward projection head reaching 35.43\%.
\begin{table}
    \centering
    \setlength{\tabcolsep}{2pt} 
    \begin{tabular}{c|c|cccc}
    \hline
        Proj. Head & Metric & $10^{-1}$ & $10^{-2}$ & $10^{-3}$ & $10^{-4}$ \\ \hline
        FFN & Unsup. Acc. & 3.69\% & 32.88\% & \textbf{35.43\%} & 29.1\% \\\hline
        TF & Unsup. Acc. &  5.05\% & 23.26\% & \textbf{40.19\% }& 32.54\% \\ \hline
    \end{tabular}
    \caption{Ablation on different \textbf{weight decay} on CIFAR100 with ResNet18 and small projection head.}
    \label{tab:weight-decay}
\end{table}

    \item \textbf{Setups for Supervised Accuracy.} To measure the supervised accuracy, we evaluated different numbers of layers and feature extraction locations including backbone and projection head. The results are shown in Table \ref{tab:linear_probing}. The results indicate that a 1-layer feed-forward network (FFN) with the backbone provides the best supervised accuracy, achieving 42.51\%, while the Transformer (TF) projection head with 1 layer achieved 46.12\%.
\begin{table}
    \centering
    \setlength{\tabcolsep}{2pt} 
    \begin{tabular}{c|c|c|c}
    \hline
        Proj. Head & Number of Layers & Backbone & Projection Head \\ \hline
        \multirow{2}{*}{FFN} 
        & 1 & \textbf{42.51\%} & 39.46\%\\
        & 3 & 39.10\% & 36.25\% \\ \hline        
        \multirow{2}{*}{TF} 
        & 1 & \textbf{46.12\%} & 44.16\%\\
        & 3 & 41.85\% & 30.34\%\\ \hline
    \end{tabular}
    \caption{Ablation on different setups to measure supervised accuracy on CIFAR100 with ResNet18 and small projection head.}
    \label{tab:linear_probing}
\end{table}
\end{itemize}

\section{Conclusion \& Limitations}

In this paper, we explored the integration of transformer models as projection heads within the contrastive learning (CL) framework, aiming to leverage the transformer's ability to capture long-range dependencies across embeddings and thereby enhance the performance of feature extraction models trained on unlabeled data. We both unveiled and explained the "Deep Fusion" phenomenon, wherein the attention mechanism of transformers progressively captures relational dependencies among samples from the same class in deeper layers.

Our experimental results demonstrate that incorporating a Transformer projection head within the SimCLR framework consistently outperforms a feed-forward network (FFN) projection head across multiple configurations, including varying backbones, batch sizes, and projection head sizes. Specifically, the Transformer projection head yields superior performance in both supervised and unsupervised tasks, with improvements ranging from 2.38\% to 5.26\% in supervised accuracy and 2.25\% to 3.20\% in unsupervised accuracy. 
Moreover, our ablation studies reveal that optimal performance is achieved with a batch size of 1024, a loss temperature of 0.2, and a weight decay rate of $10^{-3}$. Although larger projection heads offer slight accuracy gains, they also result in increased training times. 

Overall, these findings highlight the effectiveness of Transformer-based architectures in enhancing feature representation learning within contrastive learning setups, offering a promising direction for further research in this domain. In future work, we aim to address existing limitations by exploring the application of Transformers in different contexts and data modalities. Additionally, we plan to extend our theoretical analysis to further investigate the interplay between the Transformer projection head and various contrastive learning architectures.

\bibliography{aaai24}

\appendix
\onecolumn

\newpage
\section{Appendix: Algorithm to search for the bound $\rho$}
\label{algo-search-rho}
\begin{algorithm}
\caption{Algorithm to find the bound $\rho$.}
\begin{algorithmic}[1]
\REQUIRE Input features from each class $\{\mathbf{X}_1, \ldots, \mathbf{X}_\mathcal{K}\}$, with the same dimensions $d$ and number of samples $\{n_1, \ldots, n_\mathcal{K}\}$.
\REQUIRE Max iteration $T$; step size $\alpha$; convergence criterion $\tau$.
\STATE $\{\mathcal{U}_1, \ldots, \mathcal{U}_\mathcal{K}\} \gets $ basis from each class $\{\mathbf{X}_1, \ldots, \mathbf{X}_\mathcal{K}\}$

\FOR{class index $k$ in $[1,...,\mathcal{K}]$}
    \STATE $\rho_k \gets -\infty$
    \STATE $\mathbf{u}_0 \gets $ Random Initialization in $\mathbb{R}^{d}$
    \FOR{$t$ in $[0, ..., T]$ }  
        \STATE $\mathbf{u}^\perp \gets \left( \mathbf{u}_t - \mathcal{U}_k \mathcal{U}_k^\top \mathbf{u}_t \right) / \|\mathbf{u}_t - \mathcal{U}_k \mathcal{U}_k^\top \mathbf{u}_t \|_2$  \ \ \ \  // Calibrate the perpendicular vector
        \STATE $\mathbf{x}^* \gets \arg \min_{\x \notin \mathbf{X}_k} |\mathbf{x}^\top \mathbf{u}^\perp|$ \ \ \ \ \ \ \ \ \ \ \ \ \ \ \ \ \ \ \ \ \ \ \ \ \ \ \ \  // Find the minimum projection sample
        \STATE $\mathbf{u}_{t+1} \gets \left(\mathbf{u}^\perp + \alpha \cdot \mathbf{x}^* \right) / \| \mathbf{u}^\perp + \alpha \cdot \mathbf{x}^*\|_2$  \ \ \ \ \ \ \  // Move toward this sample 
        \IF{$| \rho_k  - |\mathbf{x}^{*\top} \mathbf{u}^\perp| | < \tau$}
            \STATE \textbf{break}
        \ENDIF
        \STATE $\rho_k \gets \max(\rho_k, |\mathbf{x}^{*\top} \mathbf{u}^\perp|)$ \ \ \ \ \ \ \ \ \ \ \ \ \ \ \ \ \ \ \ \ \ \ \ \ \ \ \ \  \ \ \ \ // Update the best projection value
    \ENDFOR
\ENDFOR
\STATE \textbf{return} $\min_k \rho_k$
\end{algorithmic}
\end{algorithm}


\section{Appendix: Proof of Theorem \ref{main-theorem}}
\label{pf-T2}
\begin{proof}
Our proof of Theorem \ref{main-theorem} will apply to our construction without the residual connection. The proof with residual connection follows the same logistic. Recall that
\begin{itemize}
    \item The cosine similarity between noisy and original sample is lower-bounded by a universal constant $\varepsilon \in [0,1]$:
$${\x^\top_i \tx_i}\geq (1-\varepsilon) $$
    \item When $\x_i$ is in the span of $\bu$
$$\tx_i^\top \bu^\perp \leq \sqrt{(1-(1-\varepsilon)^2)} := \delta$$
    \item When $\x_i$ is not in the span of $\bu$,
$$\tx_i^\top \bu^\perp \geq  (1-\varepsilon)\rho - \sqrt{(1-(1-\varepsilon)^2)(1-\rho^2)} := \Delta$$
\end{itemize}

Consider a matrix $\mathbf{X} \in \mathbb{R}^{n \times m}$, representing $n$ samples each in an $m$-dimensional ambient space. These samples are partitioned into $C$ clusters. Let $c_i$ denote the cluster assignment for the sample $\mathbf{x}_i$, where $c_i \in \{1,2,...,C\}$.

Introduce a weight matrix $\mathbf{W} \in \mathbb{R}^{m \times n}$, where each column $\mathbf{w}_i$ is constructed to be orthogonal to the subspaces corresponding to all clusters except the one containing $\mathbf{x}_i$.

Consequently, the product $\tX\mathbf{W}$ yields:
\[
[\tX\mathbf{W}]_{i,j} = \tx_i^\top \mathbf{w}_j = \begin{cases} \leq \delta & \text{if } c_i \neq c_j, \\
 \geq \Delta & \text{if } c_i = c_j.
\end{cases}
\]

For any pair of samples $\tx_i$, $\tx_j$ in the same cluster, the $i$-th and $j$-th rows of $\tX\mathbf{W}$, denoted as $[\tX\mathbf{W}]_{i,\cdot}$ and $[\tX\mathbf{W}]_{j,\cdot}$ respectively, have coinciding non-zero entry locations. Therefore,
\begin{align}
    [\mathbf{A}]_{i,j} = [\tX\mathbf{W}]_{i,\cdot}^\top [\tX\mathbf{W}]_{j,\cdot} \geq \nu_i \Delta^2 := \ln\alpha,
\end{align}
where $\nu_i$ is the number of samples in the same cluster as $\tx_i$.

Conversely, for any pair of samples $\tx_i$, $\tx_j$ belonging to different clusters, the non-zero entry locations in the $i$-th and $j$-th rows of $\tX\mathbf{W}$ are entirely distinct. Thus,
\begin{align}
    [\mathbf{A}]_{i,j} = [\tX\mathbf{W}]_{i,\cdot}^\top [\tX\mathbf{W}]_{j,\cdot} \leq (\nu_i + \nu_j) \delta + (n-\nu_i - \nu_j) \delta^2 := \ln \beta
\end{align}

Denote $\tx'_i$ the input of the next layer corresponding to $\tx_i$, then for the worst case, we have
$$\tx'_i = \sum_{c_j = c_i}\alpha \tx_j + \sum_{c_k \neq c_i} \beta \tx_k$$
where the normalization from softmax layer can be easily eliminated by the subsequent feed-forward layer.

Then,
\begin{itemize}
    \item When $\tx'_i$ is in the span of $\bu$,
$$\tx_j^\top \bu^\perp = \alpha \sum_{c_j = c_i} \tx_j^\top \bu^\perp + \beta \sum_{c_k \neq c_i}  \tx_k^\top \bu^\perp \leq \alpha \nu_i \delta + \beta (n-\nu_i)$$

    \item When $\tx'_i$ is not in the span of $\bu$,
$$\tx_j^\top \bu^\perp = \alpha \sum_{c_j = c_i} \tx_j^\top \bu^\perp + \beta \sum_{c_k \neq c_i}  \tx_k^\top \bu^\perp  \geq \alpha \nu_i \Delta - \beta (n-\nu_i)$$
\end{itemize}

Thus, for the affinity of next layer $\A'$
$$[\A']_{i,j} = \begin{cases}
\geq \nu_i \left( \alpha \nu_i \Delta - \beta (n-\nu_i) \right)^2 , & c_i = c_j \\
\leq (\nu_i + \nu_j)\left( \alpha \nu_i \delta + \beta (n-\nu_i) \right) + (n - \nu_i - \nu_j)\left( \alpha \nu_i \delta + \beta (n-\nu_i) \right)^2 , & c_i \neq c_j 
\end{cases}
$$

With further simplification that $1 \leq \nu_i \leq n$,
{$$[\A']_{i,j} = \begin{cases}
\geq \left( \Delta e^{\Delta^2} - n e^{n(\delta + \delta^2)} \right)^2, & c_i = c_j \\
\leq n \left(n\delta e^{n\Delta^2} + n e^{n \left(\delta + \delta^2\right)} \right) + n \left(n\delta e^{n\Delta^2} + n e^{n \left(\delta + \delta^2\right)} \right)^2
, & c_i \neq c_j 
\end{cases}
$$}

Recall Definition \ref{sharp-def} that the \textbf{sharpness} of the similarity score matrix $S$ is defined as the infimum of the ratio between the similarity scores of two points within the same cluster and the similarity scores of two points belonging to different clusters. 
$$\D(S) := \inf_{i,j,k,h} \frac{S_{i,j}}{S_{k,h}}$$
where $x_i, x_j$ are from the same cluster and $x_k, x_h$ are from different clusters.

That means that

{$$\D(S) = \frac{\alpha}{\beta} = e^{\nu_i \Delta^2 -(\nu_i + \nu_j) \delta - (n-\nu_i - \nu_j) \delta^2} \leq e^{n \Delta^2 - \delta - n \delta^2} $$}

{$$\D(S') \geq \frac{ \left( \Delta e^{\Delta^2} - n e^{n(\delta + \delta^2)} \right)^2}{
n^2 \left(\delta e^{n\Delta^2} + e^{n \left(\delta + \delta^2\right)} \right) + n^3 \left(\delta e^{n\Delta^2} +  e^{n \left(\delta + \delta^2\right)} \right)^2
}$$}

Thus, the ratio of sharpness increased by next layer is:
{$$
\frac{\D(S')}{\D(S)} \geq  \frac{ \left( \Delta e^{\Delta^2} - n e^{n(\delta + \delta^2)} \right)^2  e^{ \delta + n \delta^2 - n \Delta^2 }}{ n^2 \left(\delta e^{n\Delta^2} + e^{n \left(\delta + \delta^2\right)} \right) + n^3 \left(\delta e^{n\Delta^2} +  e^{n \left(\delta + \delta^2\right)} \right)^2} $$}

To determine the big O complexity of the ratio, we analyze both the numerator and the denominator separately and then determine the overall complexity.

The dominant term in the numerator is:

\[
\left( -n e^{n(\delta + \delta^2)} \right)^2 e^{\delta + n \delta^2 - n \Delta^2}
\\ 
= n^2 e^{2n(\delta + \delta^2)} e^{\delta + n \delta^2 - n \Delta^2} \\
= n^2 e^{n(2\delta + 3\delta^2 - \Delta^2) + \delta}
\]

The dominant term in the denominator is:

\[
n^3 e^{2n (\delta + \delta^2)}
\]

Now, combining the numerator and denominator:

\[
\gamma = O\left(\frac{n^2 e^{n(2\delta + 3\delta^2 - \Delta^2) + \delta}}{n^3 e^{2n (\delta + \delta^2)}} \right) =
O\left(\frac{1}{n} e^{n(\delta^2 - \Delta^2) + \delta}\right)
\]

This shows that the complexity heavily depends on the values of \(\delta\) and \(\Delta\). If \(\Delta^2 > \delta^2\), the exponent term will result in a decaying exponential, otherwise it could grow. However, the \( \frac{1}{n} \) term ensures a polynomial decay in complexity.

\end{proof}

\end{document}